\documentclass[lettersize,journal]{IEEEtran}
\usepackage{amsmath,amsfonts}
\usepackage{algorithm}
\usepackage{array}
\usepackage[caption=false,font=normalsize,labelfont=sf,textfont=sf]{subfig}
\usepackage{textcomp}
\usepackage{stfloats}
\usepackage{url}
\usepackage{verbatim}
\usepackage{graphicx}
\usepackage{cite}
\usepackage{algorithmicx}
\usepackage{algpseudocode}
\begin{document}

\title{Monocular 3D Lane Detection via Structure Uncertainty-Aware Network with Curve-Point Queries}

\author{Ruixin Liu, and Zejian Yuan,~\IEEEmembership{Member,~IEEE}}



\maketitle

\begin{abstract}
Monocular 3D lane detection is challenged by aleatoric uncertainty arising from inherent observation noise.
Existing methods rely on simplified geometric assumptions, such as independent point predictions or global planar modeling, failing to capture structural variations and aleatoric uncertainty in real-world scenarios.
In this paper, we propose MonoUnc, a bird's-eye view (BEV)-free 3D lane detector that explicitly models aleatoric uncertainty informed by local lane structures.
Specifically, 3D lanes are projected onto the front-view (FV) space and approximated by parametric curves. Guided by curve predictions, curve-point query embeddings are dynamically generated for lane point predictions in 3D space. 
Each segment formed by two adjacent points is modeled as a 3D Gaussian, parameterized by the local structure and uncertainty estimations. Accordingly, a novel 3D Gaussian matching loss is designed to constrain these parameters jointly.
Experiments on the ONCE-3DLanes and OpenLane datasets demonstrate that MonoUnc outperforms previous state-of-the-art (SoTA) methods across all benchmarks under stricter evaluation criteria. Additionally, we propose two comprehensive evaluation metrics for ONCE-3DLanes, calculating the average and maximum bidirectional Chamfer distances to quantify global and local errors. 
Codes are released at {https://github.com/lrx02/MonoUnc.}
\end{abstract}


\section{Introduction}
\label{sec:intro}

3D lane detection is a fundamental task in autonomous driving, identifying and localizing lanes in 3D space to support downstream tasks such as high-definition (HD) map construction~\cite{li2022hdmapnet,liao2023maptr,liu2024compact} and trajectory prediction~\cite{ma2019trafficpredict, Chen_2023_ICCV, cao2024cctr}. Monocular cameras are widely adopted for this task due to their low cost, rich visual cues, and long-range perception capabilities. 

\begin{figure}
    \centering

    \includegraphics[width=0.95\linewidth]{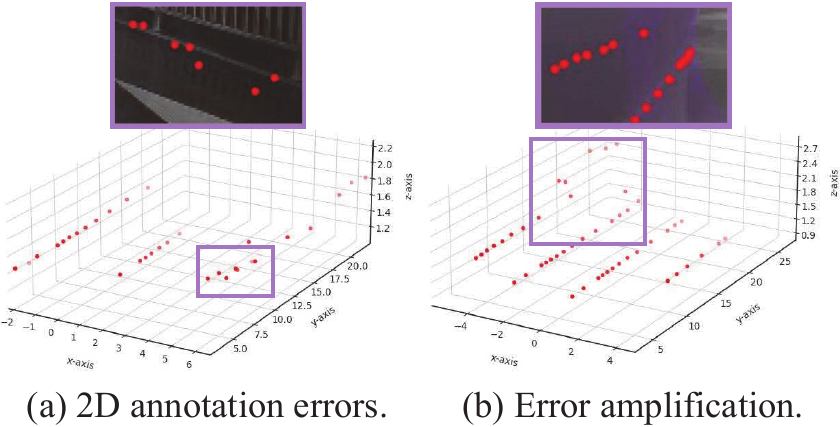}
    \caption{Aleatoric uncertainty sources in ONCE-3DLanes. (a) 2D annotation errors. (b) Error amplification.}
    \label{fig:aleatoric_uncertainty}
\end{figure}

However, monocular 3D lane detection faces aleatoric uncertainty~\cite{kendall2017uncertainties}, limiting the accuracy of recovering 3D lane structures from a single FV image.
Specifically, aleatoric uncertainty refers to inherent observation noise and randomness that cannot be reduced by collecting additional data. 
Recent monocular 3D lane datasets, such as ONCE-3Dlanes~\cite{yan2022once} and OpenLane~\cite{chen2022persformer}, generate 3D lane labels based on 2D annotations, LiDAR point clouds, and heuristic techniques, which inevitably introduce noise at multiple stages of the labeling process. Fig.~\ref{fig:aleatoric_uncertainty} (a) shows 2D annotation errors from human or automated labeling, which commonly occur under extreme lighting, weather, and occlusions. 
Fig.~\ref{fig:aleatoric_uncertainty} (b) illustrates the amplification of 2D annotation errors during 3D projection, caused by sensor calibration errors, perspective compression, and depth ambiguity. 

Despite the unavoidable aleatoric uncertainty, previous methods~\cite{garnett20193d,guo2020gen,liu2022learning, chen2022persformer, yan2022once, wang2023bev, Luo_2023_ICCV, Huang_2023_CVPR, huang2024anchor3dlane++, zheng2024pvalane} typically produce predictions without explicitly modeling it. 
Motivated by these challenges, we propose MonoUnc, an uncertainty-aware monocular lane detector, based on two key insights: 
\textit{(1) parametric curve-based representations to preserve global geometric consistency;} and
\textit{(2) local-structure-aware aleatoric uncertainty modeling to capture spatially varying noise.}

MonoUnc detects lane curves in FV space and utilizes curve-point query embeddings to localize lane points in 3D space.
Specifically, lanes are hierarchically represented: polynomial parameters approximate 2D curves, while a 3D point set captures precise details. 
Curve-point query embeddings, including curve queries extracted from FV features along predicted curves and learnable point queries, are fed into a point decoder to predict 3D lanes.

To model aleatoric uncertainty in local lane structures, we model segments formed by two adjacent points as 3D Gaussians and design a 3D Gaussian matching loss to jointly constrain their differences.
Extensive experiments are conducted on the ONCE-3DLanes and OpenLane datasets, with thorough validation of each component.
It is worth noting that existing evaluation metrics for ONCE-3DLanes rely on unilateral Chamfer distance, which fails to capture full-shape discrepancies between predictions and ground truths. To address this, we introduce two new metrics that compute the average and maximum bidirectional Chamfer distance to quantify both global and local errors. 

The main contributions of our work are threefold:
\begin{itemize}
    \item We propose MonoUnc, an end-to-end framework that predicts parametric curves to dynamically guide curve-point query generation for monocular 3D lane detection.
    \item Local lane structures with aleatoric uncertainty are modeled as 3D Gaussians, with a 3D Gaussian matching loss designed to capture spatially varying noise.
    \item Two evaluation metrics are introduced for the ONCE-3DLanes dataset, enabling a comprehensive quantification of global and local errors.
\end{itemize}

\section{Related Work}
\label{sec:related_work}

\subsection{Different Lane Representations}
Current CNN-based lane detection methods can be broadly categorized into non-parametric and parametric paradigms, depending on how lanes are represented. 

\textbf{Non-parametric paradigms} represent lanes as pixels or discrete points.
Segmentation-based methods~\cite{Pan_Shi_Luo_Wang_Tang_2018, neven2018towards, yan2022once} formulate lane detection as a pixel-level segmentation problem, requiring dense predictions and heuristic post-processing for structural modeling. 
Grid-based methods~\cite{Homayounfar_2018_CVPR, ko2021key, Liu_2023_WACV, wang2023bev} divide the representation space into grids and represent lanes as a set of points, combining sparse grid-level segmentation with point regression, where clustering techniques are typically used to separate lane instances.
In contrast to the aforementioned bottom-up methods, anchor-based methods~\cite{garnett20193d, guo2020gen, Tabelini_2021_CVPR, Zheng_2022_CVPR, chen2022persformer, Huang_2023_CVPR, huang2024anchor3dlane++} adopt a top-down framework, defining lines as anchors and regressing point offsets relative to them. 
These methods are widely used in 3D lane detection but suffer from limited adaptability due to data-specific anchor shapes.

\textbf{Parametric paradigms} represent lanes using polynomials, Bézier curves, or B-splines, which provide smooth and continuous representations without the need for post-processing.
Polynomial-based methods~\cite{Liu_2021_WACV, liu2022learning, bai2023curveformer, bai2024curveformer++} regress polynomial coefficients via sparse curve queries. While achieving higher speed, they often compromise precision due to the limited flexibility of low-degree polynomials in adapting to high-curvature lanes.
Alternatively, BézierLaneNet~\cite{Feng_2022_CVPR} estimates control points to improve curve fitting. However, adjusting any control point leads to undesirable global shape shifts due to the strong coupling among all control points. 
Unlike polynomials and Bézier curves that emphasize global adjustments, B-Splines~\cite{chen2023bsnet, Pittner_2023_WACV, Pittner_2024_CVPR} offer local control over curve segments, enabling flexible position refinements but exhibiting reduced robustness under severe occlusions.
Our MonoUnc introduces hierarchical lane representations to balance global consistency and local flexibility, utilizing parametric curves to capture global structures while accurately localizing lane points. 

\subsection{Idealized Assumptions and Uncertainty Modeling}
\textbf{Idealized assumptions} are common in monocular 3D lane detection but often unrealistic for real-world traffic scenarios.
3D-LaneNet~\cite{garnett20193d} and CLGO~\cite{liu2022learning} rely on the flat-ground assumption, predicting camera poses and employing inverse perspective mapping (IPM) to transform FV images or features into BEV space. Persformer~\cite{chen2022persformer} unifies 2D and 3D lane detection by learning robust BEV features via deformable attention based on IPM. However, these approaches struggle with non-flat roads.
LATR~\cite{Luo_2023_ICCV} instead models the ground as a dynamic plane that approximates global terrain and constrains lanes in FV space without relying on IPM. Nevertheless, it fails to capture local lane structures, especially under complex terrain variations.
These methods produce predictions based on idealized assumptions while neglecting aleatoric uncertainty, which limit the training stability and overall performance.

\textbf{Uncertainty modeling} can be divided into epistemic and aleatoric uncertainties~\cite{kendall2017uncertainties}, with the former reflecting model uncertainty and the latter capturing observation noise inherent in data.
Aleatoric uncertainty has been well studied in monocular 3D object detection~\cite{chen2020monopair, chen2021monorun, lu2021geometry, yan2024monocd}.
MonoRUn~\cite{chen2021monorun} independently models the reprojected 2D coordinates as univariate Gaussians with a robust KL loss. In contrast, GUPNet~\cite{lu2021geometry} explicitly models the dependency between 3D height and depth uncertainties using Laplace distributions, propagating height uncertainty to depth through projection and supervising with a negative log-likelihood loss.
These methods focus on object-level uncertainty while ignoring spatially varying uncertainty within local structures.
Our MonoUnc models local lane structures as 3D Gaussians to capture lateral and vertical aleatoric uncertainties relative to the local ground plane. A 3D Gaussian matching loss is designed to jointly constrain these parameters for coupled adjustments.

\section{Methodology}

\begin{figure}
    \centering
    \includegraphics[width=1\linewidth]{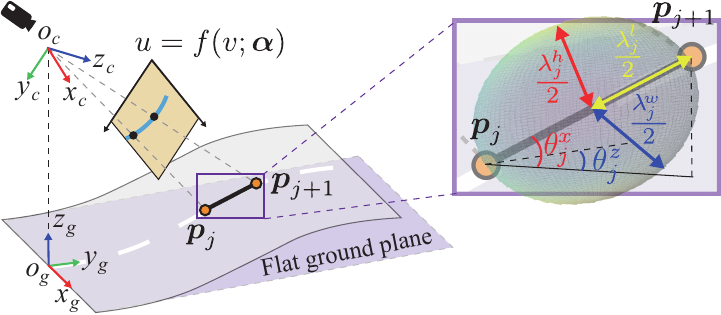}
    \caption{Illustration of coordinate systems and hierarchical lane representations. The orange lane points lie on the ground and can be projected onto the FV space, approximated by the blue curve. The black segment formed by two adjacent points $(\boldsymbol{p}_{j}, \boldsymbol{p}_{j+1})$ is represented by its length $\lambda^l_j$, uncertainties $(\lambda^w_j, \lambda^h_j)$, a pitch angle $\theta^x_{j}$, and a yaw angle $\theta^z_j$.}
    \label{fig:review}
\end{figure}

Coordinate systems relevant to lanes are illustrated in Fig.~\ref{fig:review}.
The camera coordinate system is defined with origin $o_c$ at the optical center, where the $x_c$-axis points right, the $z_c$-axis points forward, and the $y_c$-axis points downward (in meters). 
The ground coordinate system has its origin $o_g$ at the vertical projection of $o_c$ onto the ground, with $x_g$, $y_g$, and $z_g$ axes oriented right, forward, and upward, respectively.
Each 3D lane (white) is located on the ground and can be projected onto the FV space as a 2D lane (blue curve).

\subsection{Hierarchical Lane Representations}
To simultaneously capture global structures and local details of lanes, hierarchical representations are designed to consistently model parametric curves in FV space while flexibly localizing non-parametric points in 3D space.

Motivated by strong shape priors, parametric representations provide a natural and compact approximation of global lane trends. We adopt polynomial coefficients $\boldsymbol{\alpha}$, as proposed by LSTR~\cite{Liu_2021_WACV}, to represent lane curves:
\begin{equation}
    u = f(v;\boldsymbol{\alpha}),
    \label{eq:curve}
\end{equation}
where $\boldsymbol{\alpha}$ comprises shared curvature parameters $\{\rho^t\}_{t=1}^4$, individual biases $\{\beta',\beta''\}$ and boundaries $\{v^{low}, v^{up}\}$.

Alongside the curve $\boldsymbol{\alpha}$, a point set $\{\boldsymbol{p}_j\}_{j=1}^J$ of size $J$ is to capture local details, where the $j$-th point $\boldsymbol{p}_j=(x_j, y_j, z_j, vis_j)$ consists of a ground coordinate $(x_j, y_j, z_j)$ and a visibility indicator $vis_j$.
These points are distributed along predefined longitudinal $y$-coordinates $Y=\{y_j\}_{j=1}^J$.

To model local structures, we define a segment formed by two adjacent points $(\boldsymbol{p}_j, \boldsymbol{p}_{j+1})$ as an intermediate representation bridging curves and points, as shown in Fig.~\ref{fig:review}. 
Given that lane markings are thin coatings on the road surface, we assume a zero roll angle.
Each segment is characterized by its length $\lambda^l_j$, aleatoric uncertainties $(\lambda^w_j, \lambda^h_j)$, and pitch $\theta^x_j$ and yaw $\theta^z_j$ angles. Here, $\lambda^w_j$ denotes the lateral uncertainty on the local ground plane, while $\lambda^h_j$ is the vertical uncertainty orthogonal to it.

\subsection{Network Architecture}
\begin{figure*}
    \centering
    \includegraphics[width=0.9\linewidth]{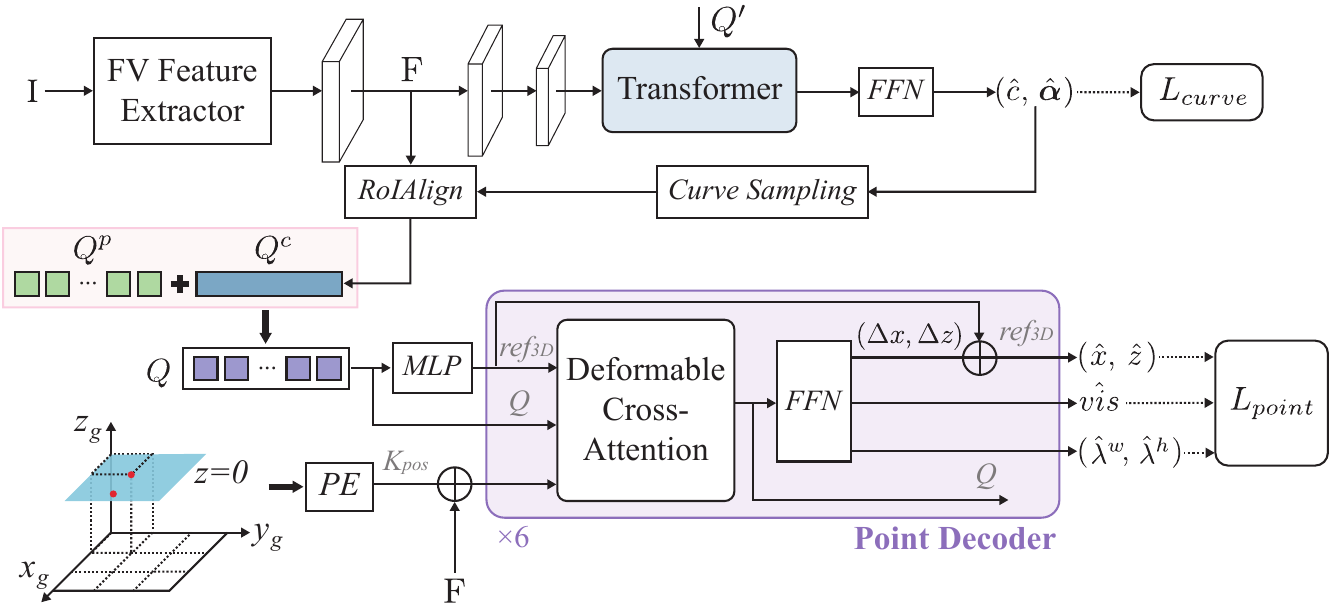}
    \caption{Overall architecture. FV feature extractor produces multi-scale features from image $\rm I$. Curve parameters are predicted by a Transformer and used to generate curve-point queries $Q$. Point decoder outputs 3D lane points with aleatoric uncertainties.}
    \label{fig:MonoUnc}
\end{figure*}

The overall architecture of MonoUnc, illustrated in Fig.~\ref{fig:MonoUnc}, consists of an FV feature extractor, a curve-point query generation module, and a point decoder for 3D lane predictions. 

\textbf{FV Feature Extractor.}
Given a monocular image ${\rm I}\in \mathbb{R}^{3\times H\times W}$ as input, a shared backbone with a $4$-layer feature pyramid network~\cite{Lin_2017_CVPR} (FPN) extracts multi-scale FV features, which are then aggregated into ${\rm F}\in \mathbb{R}^{C\times H'\times W'}$ to enhance spatial context representations.

\textbf{Curve-Point Query Generation.} To model intra-lane and inter-lane relationships, we design a curve-point query generation module that decouples lane queries into curve and point components. 
Curve queries are dynamically derived from curve predictions, with embeddings extracted from $\rm F$. 
Specifically, learnable queries $Q'\in \mathbb{R}^{K\times d'}$ are fed into a Transformer to predict $K$ sets of confidence scores and polynomial parameters, denoted as $\{(\hat{c}_k, \hat{\boldsymbol{\alpha}}_k)\}_{k=1}^K$.

\textit{(1) Curve Sampling}. Each curve is sampled at $J'$ points with uniformly spaced $v$-coordinates $\{v_{j}\}_{j=1}^{J'}$ along the image height.
Based on the $k$-th curve prediction, lane points $\{(\hat{u}_{j,k}, {v}_{j})\}_{j=1}^{J'}$ are computed via Equation~\ref{eq:curve}. For each point, a validity indicator $\hat{m}_{j,k}\in\{0,1\}$ determines whether the $j$-th point lies within the image and lane boundaries:
\begin{equation}
    \hat{m}_{j, k} = \left\{\begin{array}{ll}1, & if\ v_j\in[v^{low}_k, v^{up}_k]\land\hat{ u}_{j,k}\in[0, W)\\
    0, & otherwise\end{array}.\right.
\end{equation}

\textit{(2) RoIAlign}. Using RoIAlign~\cite{he2017mask}, features of valid points are sampled from the feature map $\rm F$ and organized as a sequence $\{\hat{f}_{j, k}\}\in \mathbb{R}^{K\times (J'\times C)}$, where:
\begin{equation}
    \hat{f}_{j, k} = \left\{\begin{array}{ll}{\rm RoIAlign}((\hat{u}_{j,k}, {v}_{j}), {\rm F}),& \hat{m}_{j, k}=1\\constants, & \hat{m}_{j, k}=0\end{array}.\right.
\end{equation}
A shared fully connected (FC) layer followed by a ReLU activation function aggregates these sequences into corresponding curve queries $Q^{c}\in \mathbb{R}^{K\times C}$:
\begin{equation}
    Q^{c}_k={\rm ReLU}({\rm FC}(\{\hat{f}_{j, k}\}_{j=1}^{J'})).
\end{equation}

Meanwhile, point-level information is encoded by learnable point queries $Q^{p}\in\mathbb{R}^{J\times C}$, which are shared across all curves and indexed with respect to the predefined $y$-coordinates $Y$. The curve-point queries $Q\in \mathbb{R}^{K\times J\times C}$ are obtained by summing:
\begin{equation}
    Q_{j,k}=Q^{c}_{j,k} + Q^{p}_{j,k}.
\end{equation}


\textbf{Point Decoder.}
A point decoder with $L$ layers is designed to iteratively refine the curve-point queries $Q$, employing deformable attention~\cite{zhu2021deformable} to improve computational efficiency.
Specifically, $Q$ is first fed into an MLP to predict 3D lane locations $(\hat{x}_{j,k}, y_j, \hat{z}_{j,k})$, where $y_j$ is indexed from the predefined $Y$-coordinates. The resulting predictions initialize the 3D reference points $ref_{3D}$, which are then projected into the FV space to serve as 2D reference points for interaction with the FV feature map $\rm F$. They are iteratively updated by the predicted 3D offsets of $\Delta x_{j,k}$ and $\Delta z_{j,k}$.
For the $l$-th decoder layer, the deformable attention process exchanges messages as follows:
\begin{equation}
\begin{split}
    Q^{(l)} &= {\rm DeformAttn}(Q^{(l-1)}, {\rm F}+K_{pos}, ref^{(l-1)}_{3D}),\\
    ref^{(l)}_{3D} &= ref^{(l-1)}_{3D} + [\Delta x,0,\Delta z]^T.
\end{split}
\end{equation}
Here, $K_{pos}\in\mathbb{R}^{C\times H'\times W'}$ denotes the 3D ground positional embeddings associated with the FV image, similar to LATR~\cite{Luo_2023_ICCV}.
These embeddings are generated by a positional encoding (PE) module, where the ground is discretized into a set of 3D points corresponding to BEV grid locations at a height of $z=0$. These points are projected into the FV space and fed into an MLP to generate $K_{pos}$.

\textbf{3D Lane Predictions.}
For each lane point, 3D location offsets $(\Delta x_{j,k}, \Delta z_{j,k})$ are predicted relative to the reference points from the last decoder layer, accompanied by a visibility indicator $\hat{vis}_{j,k}$ to determine whether the projected point is valid in the FV space.
For each segment formed by two adjacent points, two aleatoric uncertainties are predicted, consisting of a lateral uncertainty $\hat{\lambda}^w_{j,k}$ and a vertical uncertainty $\hat{\lambda}^h_{j,k}$. The final 3D lane predictions are expressed as:
\begin{equation}
\begin{split}
    \{(\hat{x}_{j,k}, y_{j}, \hat{z}_{j,k}, \hat{vis}_{j,k})|1 \leq j \leq J, 1\leq k\leq K\},\\
    \{(\hat{\lambda}^w_{j,k}, \hat{\lambda}^h_{j,k})|1 \leq j \leq J-1, 1\leq k\leq K\}.
\end{split}
\end{equation}

\subsection{Training with 3D Gaussian Matching Loss}
MonoUnc is trained with point-level and curve-level constraints. The total loss is defined as:
\begin{equation}
    L_{Total} = L_{point} + L_{curve}.
\end{equation}

\begin{figure}
    \centering
    \includegraphics[width=1\linewidth]{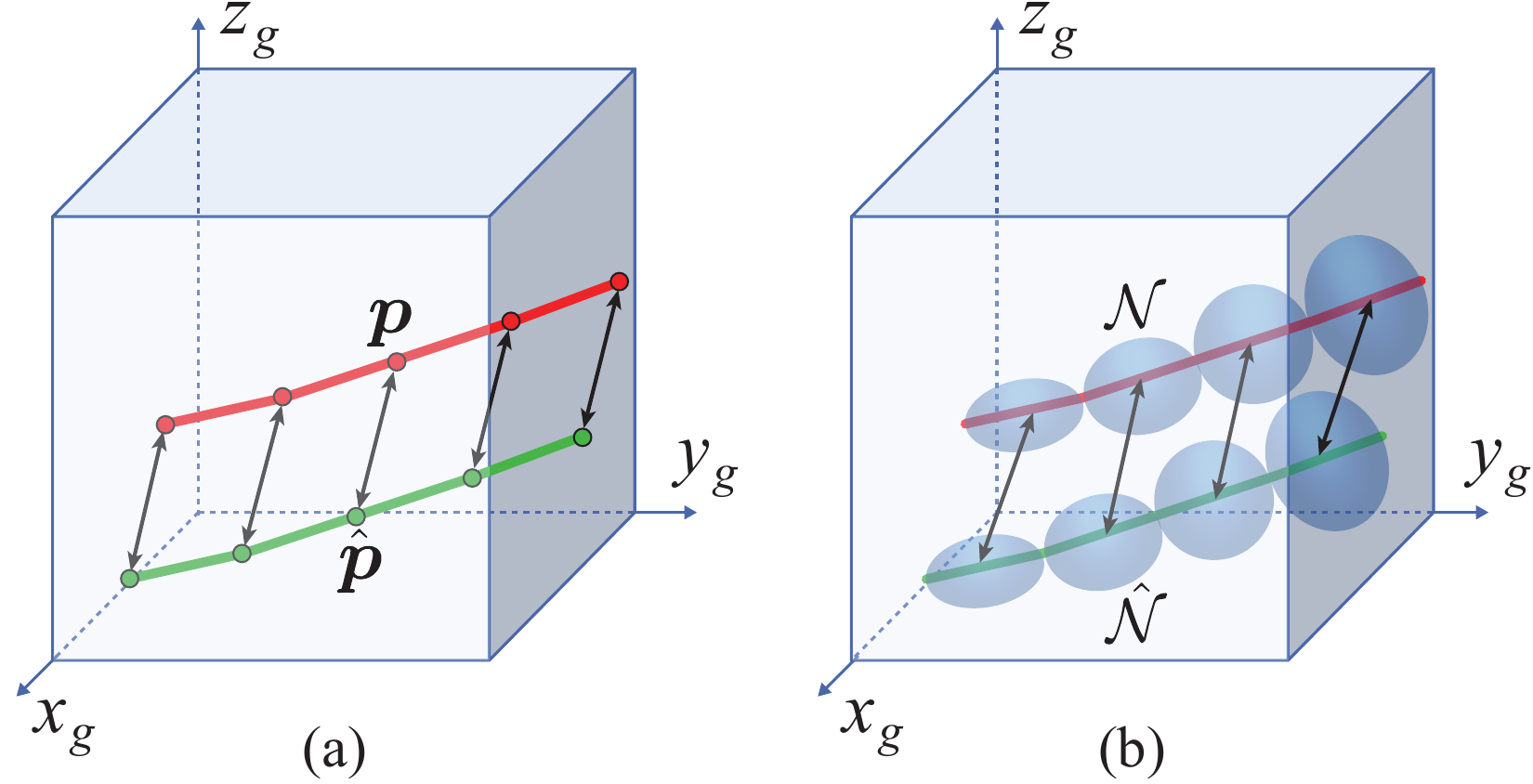}
    \caption{Components of the point-level constraint. (a) 3D location. (b) Aleatoric uncertainty.}
    \label{fig:3d_gaussian}
\end{figure}

The point-level loss $L_{point}$ supervises visibility and locations of individual points, as well as aleatoric uncertainties of segments, as shown in Fig.~\ref{fig:3d_gaussian} and defined as follows:

\begin{equation}
    L_{point}= \gamma_1 L_{unc} + L_{vis} + L_{loc},
\end{equation}
where $L_{vis}$ is the binary cross-entropy loss for visibility, and $L_{loc}$ is the L1 loss applied to the visible $x$ and $z$ coordinates:

\begin{equation}
\begin{split}
L_{loc}=&\sum_{k=1}^{K}\mathbb{I}(c_k\neq 0)\sum_{j=1}^{J}\mathbb{I}({vis}_{j,k}\neq 0)\cdot\\&(\gamma_2 |\hat{x}_{j,{\hat{\epsilon}(k)}}-{x}_{j,k}|
    +\gamma_3 |\hat{z}_{j,{\hat{\epsilon}(k)}}-{z}_{j,k}|).
\end{split}
\end{equation}
Here, $\mathbb{I}(\cdot)$ is the indicator function. Bipartite matching correspondences $\hat{\epsilon}$ are computed at the curve level using Hungarian algorithm~\cite{kuhn1955hungarian}. $\gamma_1$, $\gamma_2$, and $\gamma_3$ are weight terms.

To capture the uncertainty of local structures, we model each segment formed by two adjacent points as a 3D Gaussian $\mathcal{N}(\boldsymbol{\mu}, \boldsymbol{\Sigma})$, with mean $\boldsymbol{\mu}$ and covariance $\boldsymbol{\Sigma}$ defined as:
\begin{equation}
\begin{split}
    \boldsymbol{\mu} = (x_c, y_c, z_c)^T, \boldsymbol{\Sigma}^{1/2} =\boldsymbol{R}\boldsymbol{\Lambda} \boldsymbol{R}^T,
\end{split}
\end{equation}
where $(x_c, y_c, z_c) = (\frac{x_j+x_{j+1}}{2}, \frac{y_j+y_{j+1}}{2}, \frac{z_j+z_{j+1}}{2})$ is the 3D center of the segment, $\boldsymbol{R}$ represents the rotation matrix, and $\boldsymbol{\Lambda}$ denotes the diagonal matrix of eigenvalues.

In BEV lane detection, $\boldsymbol{R}$ and $\boldsymbol{\Lambda}$ are 2D Gaussian parameters defined by segment length, line width, and heading angle~\cite{Liu_2023_WACV, AAAI2023guan}. In contrast, 3D lane segments are modeled as 3D Gaussians (shown in Fig.~\ref{fig:review} and Fig.~\ref{fig:3d_gaussian} (b)), characterized by segment length, uncertainty, and Euler angles. The rotation matrix $\boldsymbol{R}$ is defined by the pitch angle $\theta^x = \arctan(\frac{z_{j+1}-z_{j}}{\sqrt{(x_{j+1}-x_{j})^2 + (y_{j+1}-y_{j})^2}})$ and the yaw angle $\theta^z = \arctan(\frac{y_{j+1}-y_{j}}{x_{j+1}-x_{j}})$, written as:

\begin{equation}
    R=\left(\begin{array}{ccc}
        \cos{\theta^z} & -\sin{\theta^z}\cos{\theta^x} & \sin{\theta^x}\sin{\theta^z} \\
        \sin{\theta^z} & \cos{\theta^z}\cos{\theta^x} & -\cos{\theta^z}\sin{\theta^x}\\
        0 & \sin{\theta^x} & \cos{\theta^x}\\
    \end{array}\right).
\end{equation}
And the diagonal matrix of eigenvalues $\boldsymbol{\Lambda}$ is formulated as:
\begin{equation}
    \boldsymbol{\Lambda}= 
    \left(\begin{array}{ccc}
        \frac{\lambda^l}{2} & 0 & 0 \\
        0 & \frac{\lambda^w}{2} & 0\\
        0 & 0 & \frac{\lambda^h}{2}\\
    \end{array}\right),
\end{equation}
where $\lambda^l$ denotes the segment length, computed as $\lambda^l_j=\sqrt{(x_{j+1}-x_{j})^2 + (y_{j+1}-y_{j})^2 + (z_{j+1}-z_{j})^2}$, while $\lambda^w$ and $\lambda^h$ are the lateral and vertical uncertainty estimations shared by the prediction and its corresponding ground truth.

Considering that uncertainty and local lane structure are inherently coupled rather than independent variables, we design a 3D Gaussian matching loss based on symmetric Kullback-Leibler divergence (KLD) for joint optimization:
\begin{equation}
\begin{split}
    L_{unc} = &\frac{1}{2}\sum_{k=1}^K \mathbb{I}(c_k\neq 0) \sum_{j=1}^{J-1} \mathbb{I}(vis_{j,k}\neq0 \land vis_{{j+1},k}\neq 0)\cdot\\& ({\rm KLD}(\mathcal{\hat{N}}_{j,\hat{\epsilon}(k)},\mathcal{N}_{j,k}) + {\rm KLD}(\mathcal{N}_{j,k},\mathcal{\hat{N}}_{j,\hat{\epsilon}(k)})),
\end{split}
\end{equation}
where $\mathcal{\hat{N}}_{j,\hat{\epsilon}(k)}$ and $\mathcal{N}_{j,k}$ denote the predicted and ground-truth Gaussians, respectively.

The curve-level loss $L_{curve}$ is composed of a classification loss implemented by the cross-entropy loss $L_{ce}$ and a curve fitting loss $L_{f}$ in the FV space, expressed as:
\begin{equation}
    L_{curve} = \sum_{k=1}^{K} \left(\gamma_4 L_{ce}(c_k, \hat{c}_{\hat{\epsilon}(k)}) + \mathbb{I}(c_k\neq 0) L_f(\hat{\boldsymbol{\alpha}}_{\hat{\epsilon}(k)})\right).
\end{equation}
Here, the curve fitting loss $L_{f}$ is formulated by:
\begin{equation}
\begin{split}
    L_f(\hat{\boldsymbol{\alpha}}_{\hat{\epsilon}(k)})=&\gamma_5 \sum_{j=1}^{J'}|\hat{u}_{j,{\hat{\epsilon}(k)}}-{u}_{j,k}|\\ + &\gamma_6 (|\hat{v}^{low}_{\hat{\epsilon}(k)}-v^{low}_k| + |\hat{v}^{up}_{\hat{\epsilon}(k)}-v^{up}_k|).
\end{split}
\end{equation}
The weight coefficients $\gamma_4$, $\gamma_5$, and $\gamma_6$ follow the settings in LSTR~\cite{Liu_2021_WACV}.


\section{Experiments}



\subsection{Datasets and Evaluation Metrics}
\textbf{ONCE-3DLanes}~\cite{yan2022once} is a large-scale real-world dataset built upon the ONCE dataset~\cite{mao2021one}. It contains $211$K images collected by a front-facing camera, covering diverse traffic scenarios, with accurate 3D annotations and camera intrinsic parameters provided.

\textbf{OpenLane}~\cite{chen2022persformer} is another large-scale 3D lane detection dataset derived from the Waymo dataset~\cite{sun2020scalability}. It includes $220$K frames and $880$K annotated lanes with rich geometric and semantic information. OpenLane provides both camera intrinsic and extrinsic parameters.

\textbf{ONCE-3DLanes evaluation metrics}. The official protocol computes the matching degree by calculating the Intersection over Union (IoU)~\cite{everingham2010pascal} between the BEV areas of ground-truth and predicted lanes within a predefined lane width. 
A true positive is recognized if its IoU exceeds the IoU threshold and the unilateral Chamfer distance (CD) is below the distance threshold $\tau_{CD}$. Precision (P), recall (R), F1 scores (F1), and CD errors (CDE) are reported.

\begin{figure}
    \centering
    \includegraphics[width=1\linewidth]{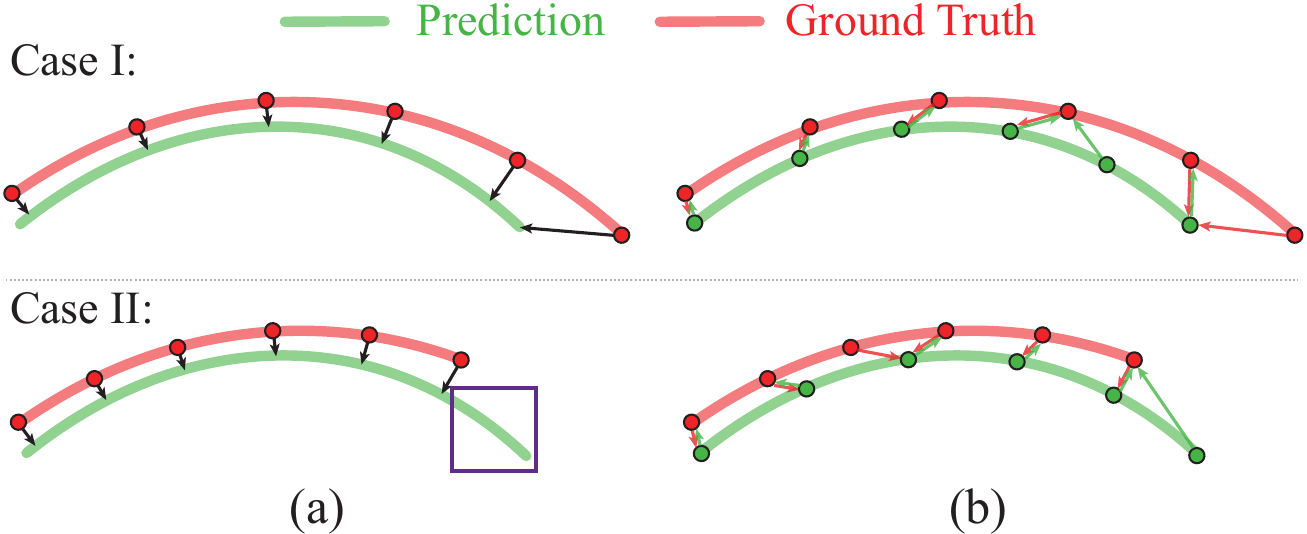}
    \caption{Illustration of unilateral and bidirectional CDs. (a) Unilateral CD. (b) Bidirectional CD.}
    \label{fig:evaluation}
\end{figure}

\begin{algorithm}
\caption{True Positives and False Positives Selection Based on Bidirectional Chamfer Distance} \label{alg:chamfer}
\begin{algorithmic}[1]
\Require  
  Ground truths $\mathcal{G} = \{G_1, \dots, G_{N_g}\}$, Predicted lanes $\mathcal{P} = \{P_1, \dots, P_{N_p}\}$, Threshold $\tau_{BCD}$, Interpolation size $N$
\Ensure  
  TP flags $\mathbf{tp} \in \{0,1\}^{N_p}$,  
  FP flags $\mathbf{fp} \in \{0,1\}^{N_p}$  

\Statex \textbf{Step 1: Dense Interpolation}  
\State Interpolate each lane to $N$ points: $\hat{G}_i, \hat{P}_j$.  
\Statex \textbf{Step 2: Bidirectional Distance Calculation}
\For{each pair $(\hat{P}_j, \hat{G}_i)$}  
  \State Compute \textbf{prediction-to-truth distance}:  
  \[
  d_{P \to G} = \frac{1}{N} \sum_{p \in \hat{P}_j} \min_{q \in \hat{G}_i} \|p - q\|_2
  \]
  \State Compute \textbf{truth-to-prediction distance}:  
  \[
  d_{G \to P} = \frac{1}{N} \sum_{q \in \hat{G}_i} \min_{p \in \hat{P}_j} \|q - p\|_2
  \]
  \State \textbf{Bidirectional distance}:  
  \[
  D_{ij} = \frac{d_{P \to G} + d_{G \to P}}{2}
  \]
\EndFor

\Statex \textbf{Step 3: TP and FP Calculation through Optimal Matching}
\State Initialize $\mathbf{tp} \gets [0]^{N_p}$, $\mathbf{fp} \gets [0]^{N_p}$, $\mathbf{covered} \gets [\text{False}]^{N_g}$
\For{each prediction $P_j$}  
  \State Find best-matched ground truth:  
  \[
  i^* = \arg\min_{i} D_{ij}
  \]
  \If{$D_{i^*j} \leq \tau_{BCD}$ \textbf{and} $\mathbf{covered}[i^*] = \text{False}$}  
    \State $\mathbf{tp}[j] \gets 1$, $\mathbf{covered}[i^*] \gets \text{True}$
  \Else  
    \State $\mathbf{fp}[j] \gets 1$  
  \EndIf  
\EndFor  

\Statex \textbf{Edge Cases}
\If{$N_g = 0$}
  \State $\mathbf{fp} \gets [1]^{N_p}$
\EndIf
\If{$N_p = 0$}
  \State \textbf{Return} $\mathbf{tp} = \emptyset$, $\mathbf{fp} = \emptyset$
\Else
\State \textbf{Return} $\mathbf{tp}$, $\mathbf{fp}$
\EndIf

\end{algorithmic}
\end{algorithm}

However, unilateral CD only accounts for errors from ground truths to predictions, failing to fully capture shape differences. As shown in Fig.~\ref{fig:evaluation} (a), it penalizes the entire predicted lane curve when the ground truth is longer than the prediction (Case I), but ignores extra false predictions (Case II, marked in purple), which leads to overly optimistic evaluations for methods producing elongated predictions.

To comprehensively assess both global and local errors, two alternative evaluation metrics based on the bidirectional Chamfer distance are introduced to quantify errors in 3D lane predictions, including the average bidirectional Chamfer distance and the maximum bidirectional Chamfer distance.
Here, the average bidirectional Chamfer distance captures the global geometric errors by jointly considering deviations in both directions between predictions and ground truths. Based on this metric, precision ${\rm P}_B$, recall ${\rm R}_B$, and F1 score ${\rm F1}_B$ are defined as follows:
\begin{equation}
\begin{split}
    {\rm P}_B&=\frac{TP}{TP+FP},\\
    {\rm R}_B&=\frac{TP}{TP+FN},\\
    {\rm F1}_B&=\frac{2\times {\rm P}_B\times {\rm R}_B}{{\rm P}_B+{\rm R}_B}.
\end{split}
\end{equation}
Here, true positives ($TP$) and false positives ($FP$) are aggregated over all validation samples. For each sample, $TP$ and $FP$ counts are determined as described in Algorithm~\ref{alg:chamfer}. The number of false negatives ($FN$) is obtained by subtracting the number of $TP$ from the total number of ground-truth lanes.
Both ground-truth and predicted lanes are densely interpolated to $N$ points, with $N$ set to $100$ in our experiments. 
The bidirectional Chamfer distance quantifies the similarity between the two unordered point sets. A predicted lane is considered a $TP$ if its average bidirectional Chamfer distance to its best-matched ground-truth lane is below the predefined threshold $\tau_{BCD}$.

The maximum bidirectional Chamfer distance (MBD) focuses on local geometric errors. Specifically, the matching degree between ground-truth lanes $\mathcal{G}=\{G_1, \cdots, G_{N_g}\}$ and predicted lanes $\mathcal{P}=\{P_1, \cdots, P_{N_p}\}$ is determined by computing the IoU, following the definition in the ONCE-3DLanes~\cite{yan2022once} dataset.
Based on the matching correspondences $\epsilon$, the maximum bidirectional Chamfer distance of the $k$-th ground truth $G_k$ and its matched prediction $P_{\epsilon(k)}$ is computed to evaluate worst-case geometric errors. Notably, MBD does not exclude overly large errors, making it a stricter and more discriminative evaluation criterion.

\textbf{OpenLane evaluation metrics}. OpenLane formulates the evaluation as a bipartite matching problem, which is solved using the minimum-cost flow, following the protocol of Gen-LaneNet~\cite{guo2020gen}. A lane prediction is considered a true positive if more than $75\%$ of its points lie within the specified distance threshold $\tau_{dist}$. 

\begin{table*}
\caption{Quantitative comparison on ONCE-3DLanes. ``P", ``R", ``F1", and ``CDE" are computed using the unilateral CD under $\tau_{CD}=0.3$ meters. ``P$_{B}$", ``R$_{B}$", ``F1$_{B}$", and ``MBD" are computed using the bidirectional CD under $\tau_{BCD}=0.3$ meters, where ``MBD" indicates the maximum bidirectional CD. All unofficial results are reported based on their best-performing configurations.}
\centering
\begin{tabular}{|c|c|c|c|c||c|c|c|c|}
\hline
Method & F1 (\%) $\uparrow$ & P (\%) $\uparrow$ & R (\%) $\uparrow$ & CDE (m) $\downarrow$ & F1$_{B} (\%) \uparrow$ & P$_{B} (\%) \uparrow$ & R$_{B} (\%) \uparrow$ & MBD (m) $\downarrow$\\
\hline

Persformer~\cite{chen2022persformer} & 74.33 & 80.30 & 69.18 & 0.074 & 26.67 & 28.94 & 24.73 & 27.268 \\
Anchor3DLane~\cite{Huang_2023_CVPR} & 74.87 & 80.85 & 69.71 & 0.060 & 33.58 & 44.61 & 26.92 & 2.305 \\
LATR~\cite{Luo_2023_ICCV} & 80.59 & 86.12 & 75.73 & 0.052 & 41.62 & 44.58 & 39.03 & 2.432 \\
MonoUnc (Ours) & \textbf{84.29} & \textbf{86.35} & \textbf{82.32} & \textbf{0.049} & \textbf{45.29} & \textbf{51.10} & \textbf{40.66} & \textbf{2.179} \\
\textit{Improvement} & $\uparrow$ \textit{3.70} & $\uparrow$ \textit{0.23} & $\uparrow$ \textit{6.59} & $\downarrow$ \textit{0.003} & $\uparrow$ \textit{3.67} & $\uparrow$ \textit{6.49} & $\uparrow$ \textit{1.63} & $\downarrow$ \textit{0.126}\\
\hline
\end{tabular}
\label{tab:comparisons_once}
\end{table*}

\begin{figure*}
    \centering
    \includegraphics[width=0.9\linewidth]{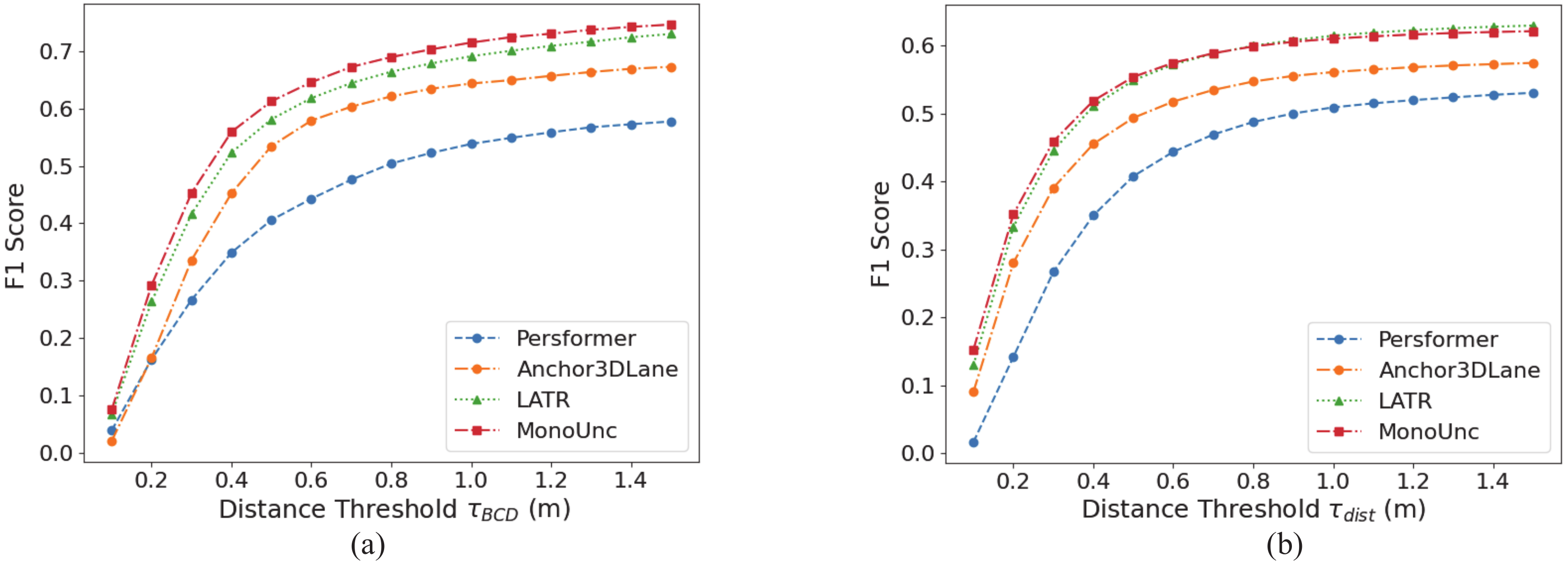}
    \caption{F1 score vs. distance threshold. (a) ONCE-3DLanes. (b) OpenLane.}
    \label{fig:comparisons_once_openlane}
\end{figure*}

\subsection{Implementation Details}
All input images are first resized to a spatial resolution of $(H, W)=(720, 960)$. The resized images are then normalized, and fed into MonoUnc. ResNet-50~\cite{He_2016_CVPR} serves as the backbone to extract $3$-scale features with spatial reduction ratios of $[\frac{1}{8}, \frac{1}{16}, \frac{1}{32}]$ relative to the input resolution.
On top of the multi-scale feature representations, a $4$-layer FPN is constructed to enhance the semantic information of features across scales, which are aggregated into a feature map $\rm F$ of size $(H', W') = (90, 120)$ and passed through the point decoder with $L=6$ layers. 
The number of curve queries and point queries is set to $K=12$ and $J=20$ for the ONCE-3DLanes dataset, and $K=40$ and $J=20$ for the OpenLane dataset to accommodate the scene and annotation complexity. 
MonoUnc is trained on $4$ Nvidia RTX 3090 GPUs with a total batch size of $32$, using the AdamW optimizer~\cite{loshchilov2018decoupled} with an initial learning rate of $2\times10^{-4}$, weight decay of $0.01$, and a cosine annealing schedule~\cite{loshchilov2017sgdr}.
Loss coefficients $\gamma_1$ to $\gamma_6$ are set to $0.5$, $2$, $10$, $3$, $5$, and $2$, respectively.

\begin{table*}
\caption{Quantitative comparison on OpenLane. $E_x$ and $E_z$ denote errors along $x$ and $z$ axes, measured in the near range ($[0, 40]$ meters) and far range ($[40, 100]$ meters). Results without * are reported in the original papers; * indicates the best-performing models provided in their official codebase.}
\label{tab:comparisons_openlane}
\centering
\setlength{\tabcolsep}{11.5mm}
\begin{tabular}{|c|c|c|c|}
\hline
Method & F1 (\%) $\uparrow$ & $E_x$(near/far) $\downarrow$ & $E_z$(near/far) $\downarrow$\\

\hline
\multicolumn{4}{|c|}{$\tau_{dist} = 0.1$ m}\\

\hline
Persformer*~\cite{chen2022persformer} & 1.7 & 0.190 / 0.120 & 0.103 / 0.076\\
Anchor3DLane*~\cite{Huang_2023_CVPR} & 9.0 & 0.109 / 0.104 & 0.061 / 0.070\\
LATR*~\cite{Luo_2023_ICCV} & 13.0 & 0.101 / 0.099 & 0.055 / 0.071\\
MonoUnc (Ours) & \textbf{15.2} & \textbf{0.100 / 0.095} & \textbf{0.055 / 0.066}\\

\hline
\multicolumn{4}{|c|}{$\tau_{dist} = 0.5$ m}\\

\hline
Persformer*~\cite{chen2022persformer} & 40.8 & 0.284 / 0.263 & 0.117 / 0.114 \\
Anchor3DLane*~\cite{Huang_2023_CVPR} & 49.4 & 0.183 / 0.203 & 0.077 / 0.102\\
LATR*~\cite{Luo_2023_ICCV} & 54.9 & 0.169 / 0.202 & 0.070 / 0.100\\
MonoUnc (Ours) & \textbf{55.4} & \textbf{0.162 / 0.187} & \textbf{0.069 / 0.092}\\

\hline
\multicolumn{4}{|c|}{$\tau_{dist} = 1.5$ m}\\

\hline
Persformer~\cite{chen2022persformer} & 50.5 & 0.485 / 0.553 & 0.364 / 0.431\\
Persformer*~\cite{chen2022persformer} & 53.1 & 0.361 / 0.328 & 0.124 / 0.129\\
Anchor3DLane~\cite{Huang_2023_CVPR} & 53.7 & 0.276 / 0.311 & 0.107 / 0.138\\
Anchor3DLane*~\cite{Huang_2023_CVPR} & 57.5 & 0.230 / 0.244 & 0.080 / 0.107\\
LATR~\cite{Luo_2023_ICCV} & 61.9 & 0.219 / 0.259 & 0.075 / 0.104\\
LATR*~\cite{Luo_2023_ICCV} & \textbf{63.0} & 0.209 / 0.250 & 0.073 / 0.105\\
MonoUnc (Ours) & 62.1 & \textbf{0.204 / 0.227} & \textbf{0.071 / 0.096}\\





\hline
\end{tabular}
\end{table*}

\begin{table*}
\caption{Quantitative comparison on OpenLane using the proposed bidirectional Chamfer distance metrics. ``P$_{B}$", ``R$_{B}$", ``F1$_{B}$", and ``MBD" are computed using the bidirectional CD under different distance thresholds $\tau_{BCD}\in\{0.1, 0.3\}$ meters, where ``MBD" indicates the maximum bidirectional Chamfer distance. * indicates the best-performing models in their official codebase.}
\centering
\setlength{\tabcolsep}{5.1mm}
\begin{tabular}{|c|c|c|c|c|c|}
\hline
Method & Backbone & F1$_{B} (\%) \uparrow$ & P$_{B} (\%) \uparrow$ & R$_{B} (\%) \uparrow$ & MBD (m) $\downarrow$ \\
\hline

\multicolumn{6}{|c|}{$\tau_{BCD}=0.1$ m}\\

\hline
Persformer*~\cite{chen2022persformer} & EfficientNet-B7~\cite{tan2019efficientnet} & 7.98 & 8.85 & 7.26 & 1.043 \\
Anchor3DLane*~\cite{Huang_2023_CVPR} & ResNet-50~\cite{He_2016_CVPR} & 24.94 & 27.74 & 22.66 & 0.952 \\
LATR*~\cite{Luo_2023_ICCV} & ResNet-50~\cite{He_2016_CVPR} & 32.57 & 34.82 & 30.59 & 0.845  \\
MonoUnc (Ours) & ResNet-50~\cite{He_2016_CVPR} & \textbf{34.13} & \textbf{37.53} & \textbf{31.30} & \textbf{0.754} \\

\hline
\multicolumn{6}{|c|}{$\tau_{BCD}=0.3$ m}\\

\hline
Persformer*~\cite{chen2022persformer} &EfficientNet-B7~\cite{tan2019efficientnet} & 48.10 & 53.36 & 43.79 & 1.043 \\
Anchor3DLane*\cite{Huang_2023_CVPR} & ResNet-50~\cite{He_2016_CVPR} & 61.19 & 68.04 & 55.59 & 0.952\\
LATR*~\cite{Luo_2023_ICCV} & ResNet-50~\cite{He_2016_CVPR} & \textbf{68.08} & 72.77 & \textbf{63.95} & 0.845\\
MonoUnc (Ours) & ResNet-50~\cite{He_2016_CVPR} & 68.07 & \textbf{74.84} & 62.42 & \textbf{0.754}\\

\hline
\end{tabular}
\label{tab:comparisons_openlane1}
\end{table*}

\subsection{Comparisons with State-of-the-Art Methods}
\textbf{Quantitative comparison on ONCE-3DLanes}. The quantitative results on the ONCE-3DLanes dataset are reported in Table~\ref{tab:comparisons_once} and visualized in Fig.~\ref{fig:comparisons_once_openlane} (a). Since the camera extrinsic parameters are not provided, we adopt the same camera configuration as used in Persformer~\cite{chen2022persformer} to ensure fair comparison and reproducibility. Under the official evaluation metrics provided by ONCE-3DLanes, our proposed MonoUnc demonstrates clear superiority over existing approaches. Specifically, MonoUnc outperforms LATR with a $3.70\%$ improvement in F1 score and a reduction of $0.003$ m in CDE, indicating more accurate geometric localization.
Furthermore, MonoUnc achieves a higher F1$_{B}$ score of $3.67\%$ than LATR under the distance threshold of $\tau_{BCD}=0.3$ m and a reduction of $0.126$ m in MBD compared with Anchor3DLane. As illustrated in Fig.~\ref{fig:comparisons_once_openlane} (a), MonoUnc consistently achieves the highest F1 scores across all evaluated distance thresholds on the ONCE-3DLanes dataset.



\textbf{Quantitative comparison on OpenLane}. The standard evaluation protocol of the OpenLane dataset adopts a distance threshold of $\tau_{dist}=1.5$ m to determine correct lane predictions, which may be insufficiently strict for safety-critical autonomous dribing scenarios, as it still tolerates notable discrepancies in both shape and localization between predictions and ground truths. Therefore, we provide a more comprehensive evaluation under a series of progressively tighter distance thresholds, as presented in Fig.~\ref{fig:comparisons_once_openlane} (b) and Table~\ref{tab:comparisons_openlane}. 
Across all thresholds, the proposed MonoUnc consistently outperforms other methods in both lateral $x$ and vertical $z$ error metrics, demonstrating more precise 3D geometry estimation. Under stricter thresholds of $\tau_{dist}=0.1$ m and $\tau_{dist}=0.5$ m, MonoUnc improves F1 scores over LATR by $2.2\%$ and $0.5\%$, respectively.
Fig.~\ref{fig:comparisons_once_openlane} (b) further illustrates that as the distance threshold decreases, the performance gap between MonoUnc and other methods becomes more pronounced, highlighting its superior capability in modeling uncertainty within local lane structures.

To further validate the effectiveness and reliability of the proposed bidirectional Chamfer distance metrics, additional quantitative experiments on the OpenLane~\cite{chen2022persformer} dataset are conducted. Table~\ref{tab:comparisons_openlane1} summarizes the results under two different distance thresholds $\tau_{BCD}\in \{0.1, 0.3\}$ m.
As the distance threshold $\tau_{BCD}$ becomes stricter, the superiority of our MonoUnc on OpenLane is further amplified, demonstrating consistency with the trends observed in the official OpenLane evaluation.




\begin{table}
\caption{Comparisons of model complexity and inference efficiency. All models are tested on a single Nvidia RTX 3090 GPU with a batch size of 1, and parameter counts are reported in millions (M).}
\centering
\setlength{\tabcolsep}{2.5mm}
\begin{tabular}{|c|c|c|c|c|}
\hline
Method & Backbone & Resolution & Params (M) & FPS \\
\hline

LATR*~\cite{Luo_2023_ICCV} & ResNet-50 & $720\times 960$ & 46.76 & 12.6  \\
MonoUnc (Ours) & ResNet-50 & $720\times 960$ & 43.29 & 13.9 \\

\hline
\end{tabular}
\label{tab:model_complexity}
\end{table}

\textbf{Model complexity and inference efficiency.}
Comparisons of model parameters and inference speed with the state-of-the-art method LATR~\cite{Luo_2023_ICCV} are presented in Table~\ref{tab:model_complexity}. Experimental results demonstrate that our proposed MonoUnc achieves an inference speed of 13.9 Frames Per Second (FPS), outperforming LATR (12.6 FPS) while using fewer parameters (43.29 M). This indicates that MonoUnc attains a more favorable trade-off between computational efficiency and model compactness, enabling real-time inference without compromising accuracy.

\begin{figure*}
    \centering
    \includegraphics[width=1\linewidth]{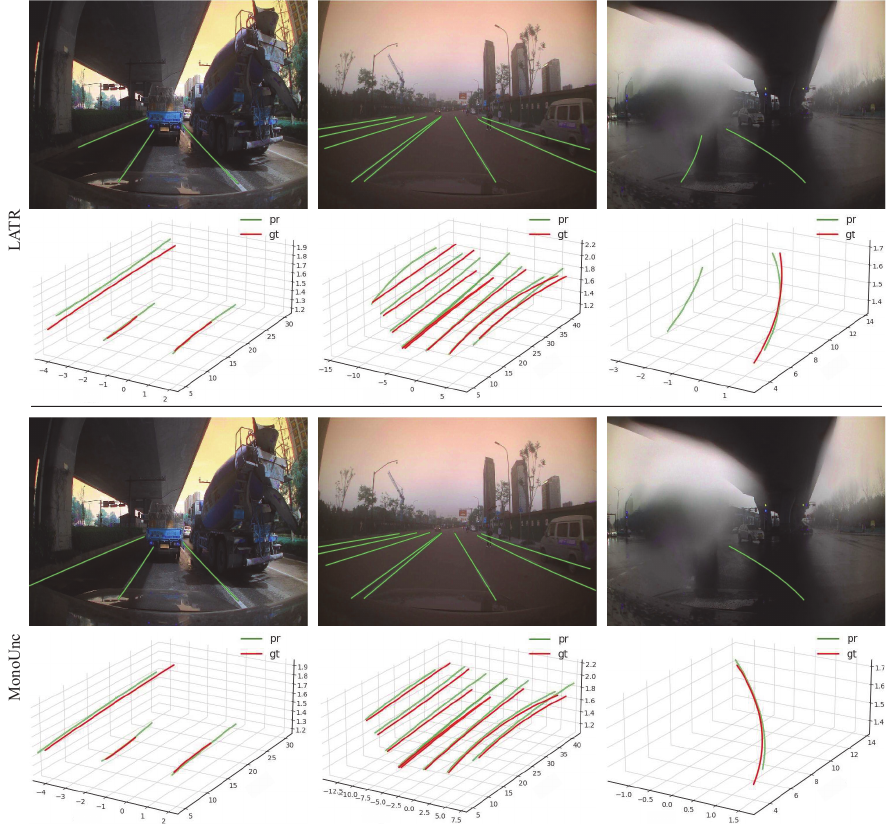}
    \caption{Qualitative comparison on ONCE-3DLanes. Ground truths and predictions are colored red and green, respectively. Best viewed in color and zoomed in for details.}
    \label{fig:comparisons_vis}
\end{figure*}

\textbf{Qualitative comparison on ONCE-3DLanes}. Fig.~\ref{fig:comparisons_vis} illustrates a qualitative comparison between LATR (upper row) and MonoUnc (lower row) on the ONCE-3DLanes dataset. Although both methods produce visually comparable detection results in the front-view images, MonoUnc generates more accurate and geometrically consistent 3D predictions across various scenarios, including curves, occlusions, and distant lanes. Additional qualitative results, covering challenging cases such as rainy conditions, heavy occlusions, uphill/downhill roads, fork/merge scenes, and curved lanes, are provided in the supplementary materials.

\subsection{Ablation Studies}
We conduct ablation studies on the ONCE-3DLanes dataset to systematically evaluate the contributions of different components in our methods, including the impact of different query embeddings, the effectiveness of the proposed 3D Gaussian matching loss, and the design choices related to uncertainty modeling.

\begin{figure*}
    \centering
    \includegraphics[width=0.9\linewidth]{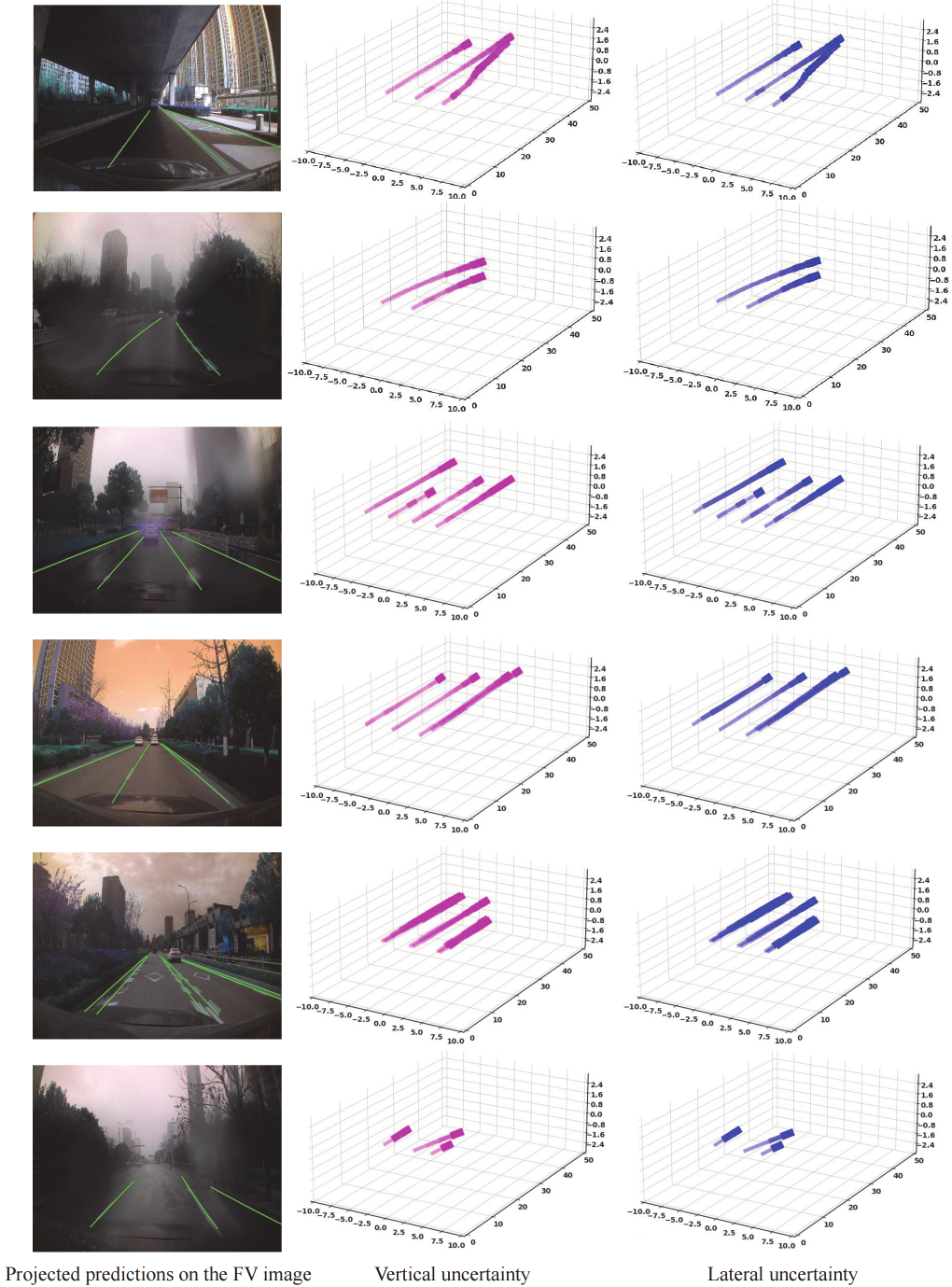}
    \caption{Visualization of intermediate results. Vertical and lateral uncertainties are shown in pink and blue, respectively. Thicker lane segments with higher opacity indicate greater uncertainty.}
    \label{fig:intermediate}
\end{figure*}

\textbf{Different query embeddings.}
Table~\ref{tab:components} presents a comparison of different query embeddings. The combination of ``$Q^{c}$ + $Q^{p}$" demonstrates superior performance over ``IAM + $Q^{p}$", achieving an improvement of $2.28\%$ in F1$_{B}$ and a reduction of $0.221$ m in MBD. Here, ``IAM" refers to the lane-level queries generated by the instance activation map~\cite{cheng2022sparse}. The reason is that $Q^{c}$ is capable of providing a parametric shape prior for lanes, which helps the model better capture the overall geometric structure and curvature of 3D lanes. 

\begin{table}
\caption{Ablation studies on query embeddings and uncertainty modeling.}

\centering
\begin{tabular}{|c|c|c|c|c|c|c|}
\hline
$Q$ & {$L_{unc}$} & {F1$_{B} (\%)$} & {P$_{B} (\%) $} & {R$_{B} (\%) $} & {MBD (m) }\\

\hline

IAM+$Q^{p}$ & - & 41.62 & 44.58 & 39.03 & 2.432 \\
$Q^{c}$+$Q^{p}$ & - & 43.90 & 49.37 & 39.52 & 2.211 \\
$Q^{c}$+$Q^{p}$ & $\checkmark$ & \textbf{45.29} & \textbf{51.10} & \textbf{40.66} & \textbf{2.179} \\

\hline
\end{tabular}\label{tab:components}
\end{table}

\textbf{3D Gaussian matching loss.}
We further analyze the effect of the 3D Gaussian matching loss $L_{unc}$ by varying its weight $\gamma_1$, as reported in Tables~\ref{tab:components} and~\ref{tab:unc_loss}. Compared to setting $\gamma_1$ to $0$, assigning an appropriate weight of $\gamma_1 = 0.5$ improves F1$_{B}$ by $1.39\%$. However, increasing $\gamma_1$ to 1 overemphasizes uncertainty, leading to a decline in 3D point localization accuracy. These observations highlight the importance of balancing the contribution of the 3D Gaussian matching loss to achieve optimal performance in 3D lane detection.

\begin{table}
\caption{Ablation studies on the 3D Gaussian matching loss.}
\centering
\setlength{\tabcolsep}{3.8mm}
\begin{tabular}{|c|c|c|c|c|}
\hline
$\gamma_{1}$ & F1$_{B} (\%)$ & P$_{B} (\%)$ & R$_{B} (\%)$ & MBD (m) \\
\hline
0 & 43.90 & 49.37 & 39.52 & 2.211 \\
0.25 & 44.45 & 51.65 & 39.01 & \textbf{2.051} \\
0.5 & \textbf{45.29} & \textbf{51.10} & 40.66 & 2.179 \\
0.75 & 45.19 & 49.87 & \textbf{41.32} & 2.150 \\
1 & 41.59 & 47.62 & 36.91 & 2.381 \\
\hline
\end{tabular}
\label{tab:unc_loss}
\end{table}

\begin{figure}
    \centering
    \includegraphics[width=1\linewidth]{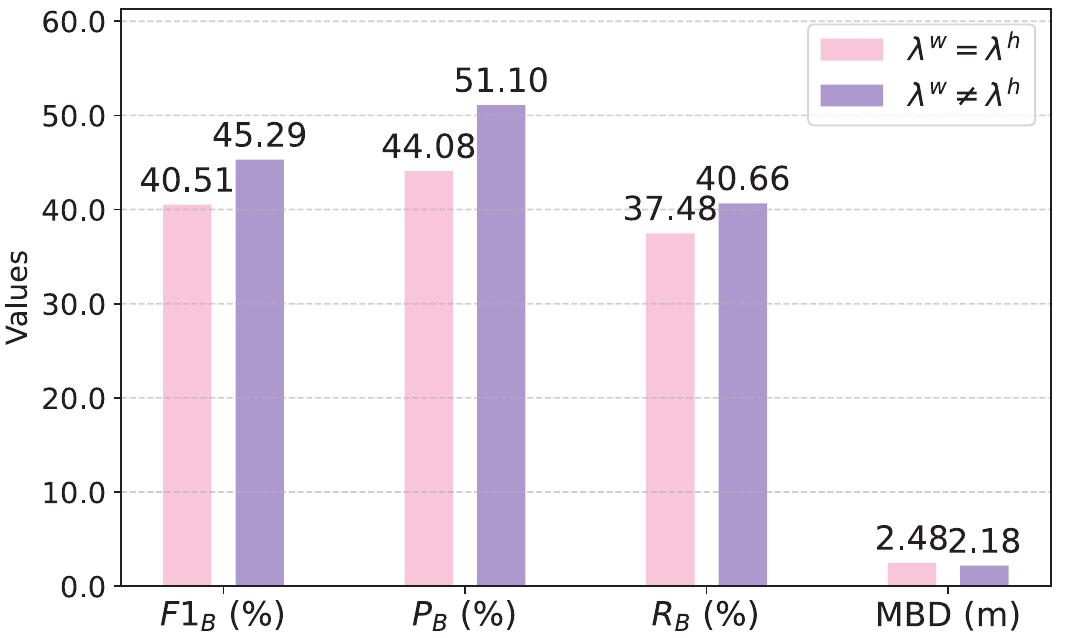}
    \caption{Ablation studies on uncertainty modeling.}
    \label{fig:uncertainty_ablation}
\end{figure}

\textbf{Uncertainty modeling choices}.
To clarify our choice of modeling aleatoric uncertainty in local lane structures, which includes lateral $\lambda^w$ and vertical $\lambda^h$ uncertainties, we compare shared uncertainty $\lambda^w = \lambda^h$ and separate uncertainty $\lambda^w\neq \lambda^h$.
As shown in Fig.~\ref{fig:uncertainty_ablation}, using shared uncertainty results in a noticeable performance drop of $4.78\%$ in F1$_{B}$ compared to separate uncertainty. This approach even performs worse than the baseline without uncertainty modeling ($40.51\%$ \textit{vs.} $43.90\%$ in F1$_{B}$ and $2.482$ m \textit{vs.} $2.211$ m in MBD). The reason is that lateral and vertical directions exhibit different observation noise characteristics, requiring separate modeling.
Fig.~\ref{fig:intermediate} visualizes intermediate results, depicting lanes as a series of segments, where thicker segments with higher opacity indicate greater uncertainty. Both lateral and vertical uncertainties increase with depth. 

\section{Conclusion}

In this work, we propose MonoUnc, a Transformer-based framework for monocular 3D lane detection. By leveraging curve-point queries, MonoUnc effectively captures the multi-scale geometric structures of lanes, enabling accurate 3D localization. Furthermore, local-structure-aware aleatoric uncertainty is modeled as a 3D Gaussian and optimized via a 3D Gaussian matching loss to learn spatially varying noise. 
Beyond lane detection, our method shows promise for applications such as camera-based 3D object detection and 3D semantic scene completion. In future work, multi-modal and multi-frame data will be incorporated to enhance the robustness and scalability.

\bibliographystyle{IEEEtran}
\bibliography{reference}

\begin{IEEEbiography}[{\includegraphics[width=1in,height=1.25in,clip,keepaspectratio]{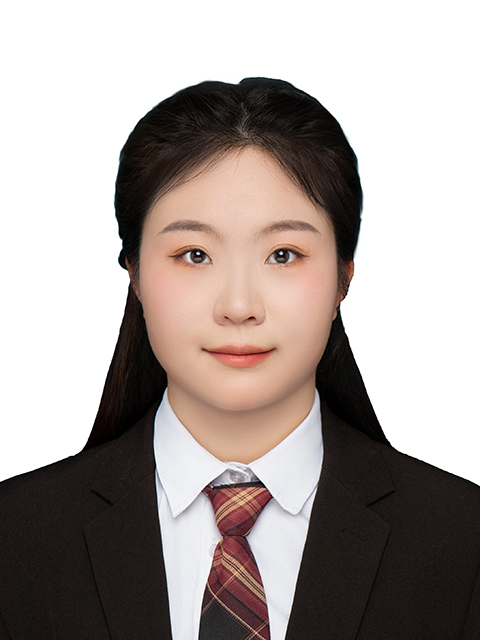}}]{Ruixin Liu}
received the B.S. degree in computer science and technology from Tianjin University, Tianjin, China, in 2019. She is currently pursuing the Ph.D. degree in control science and engineering under the supervision of Dr. Zejian Yuan in Xi'an Jiaotong University, Xi'an, China. Her research interests include computer vision and deep learning.
\end{IEEEbiography}

\vspace{11pt}
\vspace{-33pt}
\begin{IEEEbiography}
[{\includegraphics[width=1in,height=1.25in,clip,keepaspectratio]{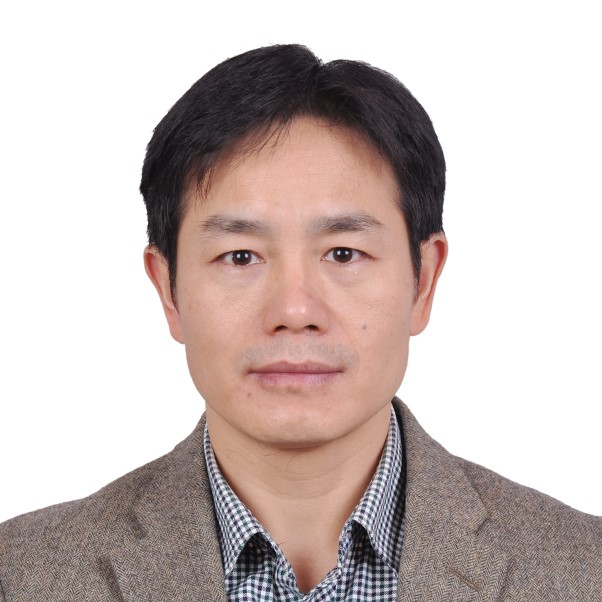}}]{Zejian Yuan}
(Member, IEEE) received the M.S.
degree in electronic engineering from the Xi’an University of Technology, Xi’an, China, in 1999, and the Ph.D. degree in pattern recognition and intelligent systems from Xi’an Jiaotong University,
China, in 2003. He is currently a Professor with the College of Artificial Intelligence, Xi’an Jiaotong University, and a member of the Chinese Association of Robotics. His research interests include image processing, pattern recognition, and machine learning in computer vision.
\end{IEEEbiography}

\vfill

\end{document}